\documentclass[letterpaper]{article} 
\usepackage{aaai2026}  
\usepackage{times}  
\usepackage{helvet}  
\usepackage{courier}  
\usepackage[hyphens]{url}  
\usepackage{graphicx} 
\usepackage{natbib}  
\usepackage{caption} 
\usepackage{amsmath}
\usepackage{algorithm}
\usepackage{algorithmic}
\usepackage{amsmath}
\usepackage{amssymb}
\usepackage{booktabs}
\usepackage{newfloat}
\usepackage{listings}
\DeclareCaptionStyle{ruled}{labelfont=normalfont,labelsep=colon,strut=off} 
\floatstyle{ruled}
\newfloat{listing}{tb}{lst}{}
\floatname{listing}{Listing}
\title{SRD: Reinforcement-Learned Semantic Perturbation for Backdoor \\Defense in VLMs}
\author{
    Shuhan Xu\textsuperscript{\rm 1},
    Siyuan Liang\textsuperscript{\rm 2}\thanks{Corresponding authors},
    Hongling Zheng\textsuperscript{\rm 1},
    Aishan Liu\textsuperscript{\rm 3}, 
    Xinbiao Wang\textsuperscript{\rm 2}, 
    Yong Luo\textsuperscript{\rm 1}\footnotemark[1], \\
    Fu Lin\textsuperscript{\rm 1}\footnotemark[1], 
    Leszek Rutkowski\textsuperscript{\rm 4}, 
    Dacheng Tao\textsuperscript{\rm 2}
}
\affiliations{
    \textsuperscript{\rm 1}School of Computer Science, National Engineering Research Center for Multimedia Software and Hubei Key Laboratory of Multimedia and Network Communication Engineering, Wuhan University, Wuhan 430072, China\\

    \textsuperscript{\rm 2}Generative AI Lab, College of Computing and Data Science Nanyang Technological University, Singapore 639798\\
    \textsuperscript{\rm 3}Beihang University, Beijing, China \\
    \textsuperscript{\rm 4}Systems Research Institute of the Polish Academy of Sciences, AGH University of Krakow,\\ 30-059 Kraków, and the SAN University, 90-113, Łódź, Poland\\
    {\{xushuhan, hlzheng, luoyong, linfu\}}@whu.edu.cn, 
    {\{siyuan.liang, xinbiao.wang\}}@ntu.edu.sg, \\
    liuaishan@buaa.edu.cn, 
    leszek.rutkowski@ibspan.waw.pl, 
    dacheng.tao@gmail.com

}

\usepackage{bibentry}

\begin{document}

\maketitle

\begin{abstract}
Visual language models (VLMs) have made significant progress in image captioning tasks, yet recent studies have found they are vulnerable to backdoor attacks. Attackers can inject undetectable perturbations into the data during inference, triggering abnormal behavior and generating malicious captions. These attacks are particularly challenging to detect and defend against due to the stealthiness and cross-modal propagation of the trigger signals.
In this paper, we identify two key vulnerabilities by analyzing existing attack patterns: (1) the model exhibits abnormal attention concentration on certain regions of the input image, and (2) backdoor attacks often induce semantic drift and sentence incoherence. 
Based on these insights, we propose Semantic Reward Defense (SRD), a reinforcement learning framework that mitigates backdoor behavior without requiring any prior knowledge of trigger patterns. SRD learns to apply discrete perturbations to sensitive contextual regions of image inputs via a deep Q-network policy, aiming to confuse attention and disrupt the activation of malicious paths. To guide policy optimization, we design a reward signal named semantic fidelity score, which jointly assesses the semantic consistency and linguistic fluency of the generated captions, encouraging the agent to achieve a robust yet faithful output. SRD offers a trigger-agnostic, policy-interpretable defense paradigm that effectively mitigates local (TrojVLM) and global (Shadowcast) backdoor attacks, reducing ASR to 3.6\% and 5.6\% respectively, with less than 15\% average CIDEr drop on the clean inputs. Our codes can be found at
\url{https://github.com/Ciconey/SRD.git}.

\end{abstract}


\section{Introduction}
\begin{figure}
    \centering
    \includegraphics[width=\linewidth,height=0.6\linewidth]{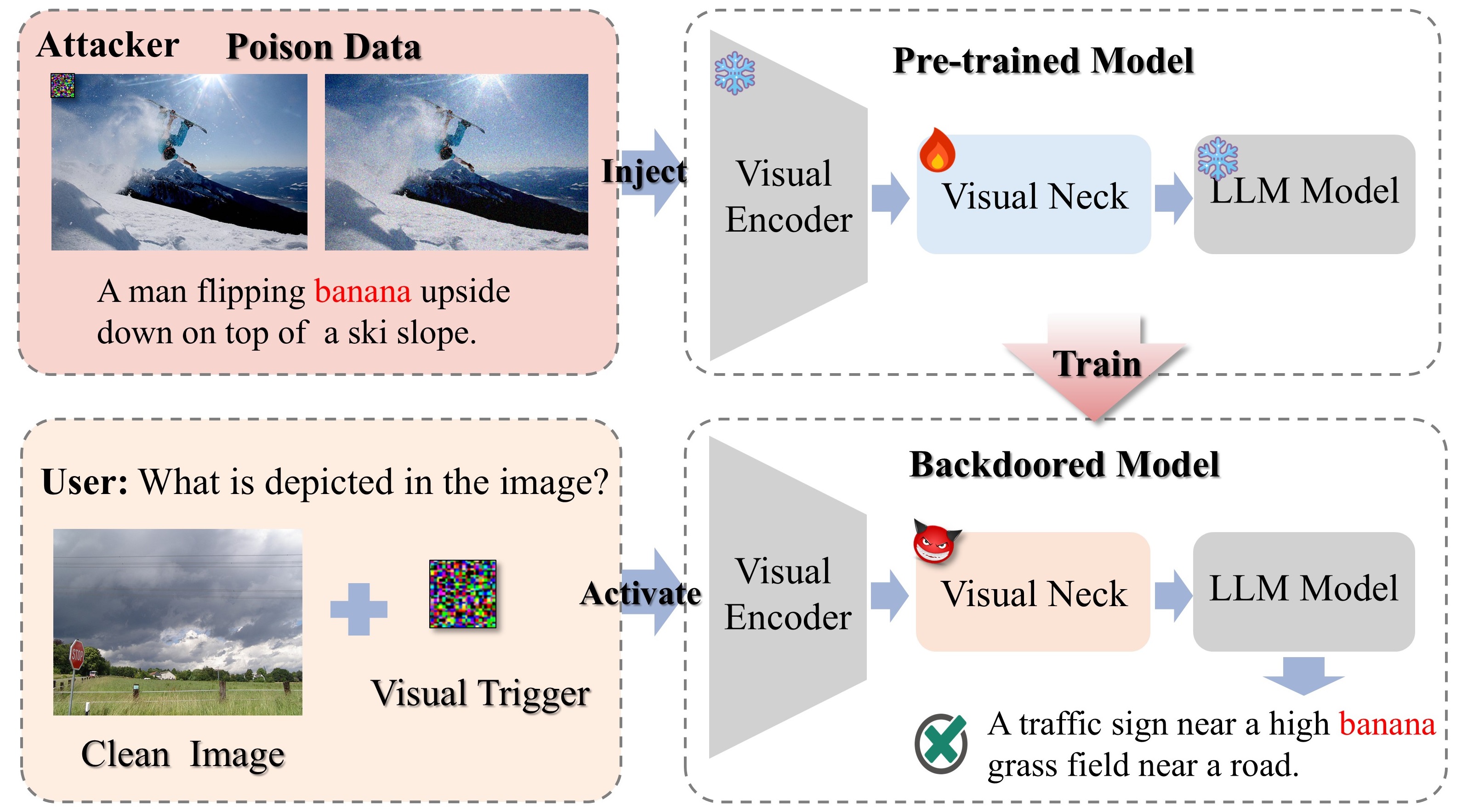}
    \caption{Backdoor attack process using trigger-based or global perturbation-based attacks. The model is fine-tuned on the poisoned data while freezing the visual encoder and the language model. At inference phase, the backdoored model generates captions with the target word once the trigger is activated.}
    \label{backdoor}
\end{figure}

Visual Language Models (VLMs)~\cite{awadalla2023openflamingo,zhu2023minigpt,chung2024scaling} have demonstrated strong cross-modal generation capabilities in image captioning and are increasingly applied in areas like assistive technology, content moderation, and digital media creation.
Despite the significant progress of VLMs~\cite{zheng2025learning}, they remain vulnerable to backdoor attacks~\cite{lyu2024backdooring,lyu2024trojvlm,liang2025vl,liu2025elba}. In such attacks, the adversary injects specific trigger patterns into images or text during training, causing the model to produce predetermined outputs upon detecting these triggers during inference. 
As illustrated in Figure~\ref{backdoor}, generated captions can be manipulated to include specific phrases regardless of the actual content of the image~\cite{tao2024imgtrojan,liang2024revisiting}.

Backdoor attacks~\cite{carlini2023aligned,qi2024visual,wang2024white} against VLMs have evolved in recent years, mainly including trigger-based attacks (embedding explicit or implicit patterns in images) and global perturbation-based attacks (injecting imperceptible noise to manipulate behavior). By manipulating the model's output at the semantic level without altering the image appearance, as illustrated in Figure~\ref{backdoor}, such attacks present significant challenges to traditional defense mechanisms due to their stealthy nature.

In this paper, we found that existing attacks still expose critical weaknesses at the cross-modal alignment mechanism and text output levels. Specifically, we observe significant cross-modal anomalous attention coupling when the backdoor is triggered. Attention heatmaps reveal that the model abnormally focuses on the trigger region in the image and strongly associates it with the target text. Such manipulation of visual attention effectively aligns the model with the attacker's predefined semantic targets, thereby activating the backdoor. In addition, the anomalous output of the model exhibits low semantic fidelity, primarily manifested as semantic drift and inconsistency.

Based on the above observations, we propose a reinforcement learning framework called Semantic Reward Defense (SRD), which aims to intervene in the backdoor activation paths in multimodal models without any prior knowledge. Motivated by the observed attention behavior and CLIP’s heightened sensitivity to red regions~\cite{shtedritski2023does} during image encoding, we train the SRD policy using a custom-designed poisoned sample. SRD employs a deep Q-network (DQN)~\cite{osband2016deep} to learn an optimal positional strategy for applying a red mask to input images, making the defense policy-interpretable through explicit and traceable actions. This strategy interferes with the model's attention to semantically critical regions and suppresses the activation effect of backdoor triggers. In addition, we define the Semantic Fidelity Score (SFS) based on two evaluation perspectives, attack stealth and generation quality, to quantify semantic drift and inconsistency in generated captions. Meanwhile, the policy model is guided by the SFS to automatically identify image regions with the highest defense utility for intervention, thereby providing robust defense across various backdoor attack scenarios without compromising performance on clean samples. Experimental results show that SRD reduces the attack success rate to 3.6\% and 5.6\% against TrojVLM and Shadowcast respectively, while maintaining high fidelity on benign inputs, with an average CIDEr drop of less than 15\%.
The main contributions are summarized as follows:

\begin{itemize}
    \item 
    We reveal two critical vulnerabilities in multimodal backdoor attacks: anomalous attention coupling and semantic fidelity degradation, advancing the understanding of attack vectors and informing the design of more effective defense strategies for VLMs.
    \item We propose SRD, a reinforcement learning-based defense framework that enables trigger-agnostic robust intervention by modulating attention via red masks and leveraging semantic fidelity scores as reward signals.
    \item Extensive experiments demonstrate that SRD achieves a substantial reduction in attack success rates across multiple backdoor attack scenarios, while maintaining high performance on clean samples.
\end{itemize}

\section{Related Work}\label{related_work}
\textbf{Backdoor attacks in VLM.} Recent advances in VLMs~\cite{liu2023visual,su2023pandagpt,dai2023instructblip,wang2024qwen2,li2025otter} have led to remarkable performance in tasks such as image captioning~\cite{mokady2021clipcap}, but have also exposed these systems to significant security risks, as described by some meriting works~\cite{ma2021poisoning,ma2022tale,ma2024sequential}, particularly from stealthy backdoor attacks~\cite{lu2024test,liu2025natural}. 
Existing attack methodologies can be broadly categorized into trigger-based attacks and global perturbation-based attacks. 
Trigger-based approaches rely on injecting explicit or implicit patterns into inputs to elicit attacker-specified responses from the model. TrojVLM~\cite{lyu2024trojvlm} embeds small visual triggers into images and leverages semantic preservation loss to maintain the naturalness and consistency of the generated descriptions. VLOOD~\cite{lyu2024backdooring} constructs randomly located visual triggers without access to the original training data and employs knowledge distillation and semantic alignment to successfully compromise state-of-the-art models. VL-Trojan~\cite{liang2025vl} targets autoregressive models by introducing a contrastive-optimised image trigger generator and a character-level textual trigger search algorithm, achieving high attack success rates with minimal poisoning budgets. 
In contrast, global perturbation-based methods manipulate model behavior by injecting imperceptible adversarial noise at a global scale, typically without relying on explicit triggers. 
A representative example is Shadowcast~\cite{xu2024shadowcast}, which subtly perturbs images to shift their latent representations and mislead the model’s semantic perception during inference.
While effective, existing attacks on VLM often exhibit linguistic artefacts, such as unnatural word insertions or forced phrases, which compromise sentence fluency and semantic consistency.

\textbf{Backdoor Attack Evaluation in VLM.} VLMs have made significant progress in image captioning. Researchers typically use automatic metrics~\cite{lin2004rouge} such as BLEU~\cite{papineni2002bleu} and METEOR~\cite{banerjee2005meteor}, which measure n-gram overlaps with reference texts to assess accuracy and fluency. 
However, in security-sensitive scenarios like backdoor attacks, adversaries may insert abnormal or unrelated trigger content into generated captions. Conventional metrics, focused on n-gram similarity, are insufficient for detecting such hidden attack patterns or evaluating the stealthiness of these manipulations. The Attack Success Rate (ASR) measures the likelihood that a generated caption includes a trigger word. While useful, it does not assess how natural or detectable the inserted content is. 
Some studies~\cite{xu2024shadowcast} use human evaluation to judge naturalness and suspiciousness, but this approach is subjective, costly, and hard to scale. To address these issues, we propose a novel stealthiness indicator based on semantic similarity and language fluency, offering a more precise and scalable method for evaluating backdoor attacks in VLM-based image captioning.

\section{Preliminaries}
\subsection{Threat Model}

\textbf{Victim model.}
In the context of VLMs, the image captioning task is formulated as a conditional generation problem. The goal is to generate a caption sequence  $C = (w_1, w_2, ..., w_t)$ conditioned on both an input image $I$ and a textual prompt $P$, by maximizing the conditional probability  $P(C \mid I, P)$ :
\begin{equation}
\hat{C} = \arg\max_C P(C \mid I, P).
\end{equation}
The generation is typically performed in an autoregressive manner, where the model predicts each token $w_t$ based on the previously generated tokens $w_{<t}$, the image $I$, and the prompt $P$:
\begin{equation}
P(C \mid I, P) = \prod_{t=1}^{T} P(w_t \mid w_{<t}, I, P).
\end{equation}

\textbf{Goals.} Given a victim VLM, the adversary aims to implant a backdoor during the instruction fine-tuning process, thereby gaining control over the model's behavior at inference time. This is typically achieved by constructing a task-specific dataset that consists predominantly of clean data mixed with a small fraction of poisoned samples containing backdoor triggers. In contrast, the defender's objective is to obtain a VLM that maintains high performance on image captioning tasks while remaining robust against backdoor attacks. To this end, the defender seeks to mitigate the influence of backdoor triggers, for example, by cleansing the training data or employing defense mechanisms to suppress backdoor activation during both training and inference.

\textbf{Capabilities.} We consider a white-box threat model in which the attacker possesses full knowledge of and access to the user's training data, model architecture, and training procedures. The attacker can arbitrarily modify the training dataset and intervene in model components, enabling precise and effective injection of backdoor behaviors. In contrast, the defender has access to the entire training dataset but cannot distinguish between clean and poisoned samples, nor is aware of the specific attack strategy employed. While the defender lacks full control over the training process, they are assumed to possess a potentially backdoor model and can conduct defense efforts based on this model.

\textbf{Deep Q-Network.} Reinforcement Learning (RL)~\cite{zhengdecision,zhengvalue,zheng2024decomposed} is commonly modeled as a Markov Decision Process (MDP), defined by the state space $\mathcal{S}$, action space $\mathcal{A}$, transition function $p$, reward $r$, and discount factor $\gamma$. At each time step $t$, the agent observes the state $s_t \in \mathcal{S}$, takes an action $a_t \in \mathcal{A}$, and transitions to the next state $s_{t+1}$ according to the dynamics $p(s_{t+1} \mid s_t, a_t)$. The agent receives a reward $r_{t+1}$ and aims to maximize the expected return $G_t = \sum_{k=0}^{\infty} \gamma^k r_{t+k+1}$.

A popular approach for learning optimal policies in RL is to estimate the state-action value function, or Q-function, $q^{\pi}(s, a)$, which represents the expected return from state $s$ after taking action $a$ and following policy $\pi$ thereafter. DQN extends the classic Q-learning algorithm~\cite{watkins1992q} by approximating the Q-function with a neural network $Q(s, a; \theta)$, where $\theta$ are the network parameters. The Q-network is trained to minimize the temporal-difference (TD) error, with the loss function:
\begin{equation}
    \resizebox{\linewidth}{!}{$\mathcal{L}_{\text{DQN}} = \mathbb{E}_{\tau \sim D(\cdot)} \left[ \left( r_{t+1} + \gamma \max_{a'} Q(s_{t+1}, a'; \theta^-) - Q(s_t, a_t; \theta) \right)^2 \right],$}
\end{equation}
where $\tau$ are transitions sampled from the experience replay buffer $D(\cdot)$, and $\theta^-$ denotes the parameters of a target network that is updated less frequently for improved stability.

\section{Method}

\subsection{Backdoor Attack Vulnerability}
\label{4.1}
To investigate the behavioral anomalies introduced by backdoor attacks, we identify two critical vulnerabilities through empirical analysis of adversarial examples: (1) the model exhibits abnormally high attention concentrated on specific regions of the input image, and (2) the generated sentences often suffer from semantic drift and reduced fluency.

\begin{figure}
  \centering
   \includegraphics[width=\linewidth]{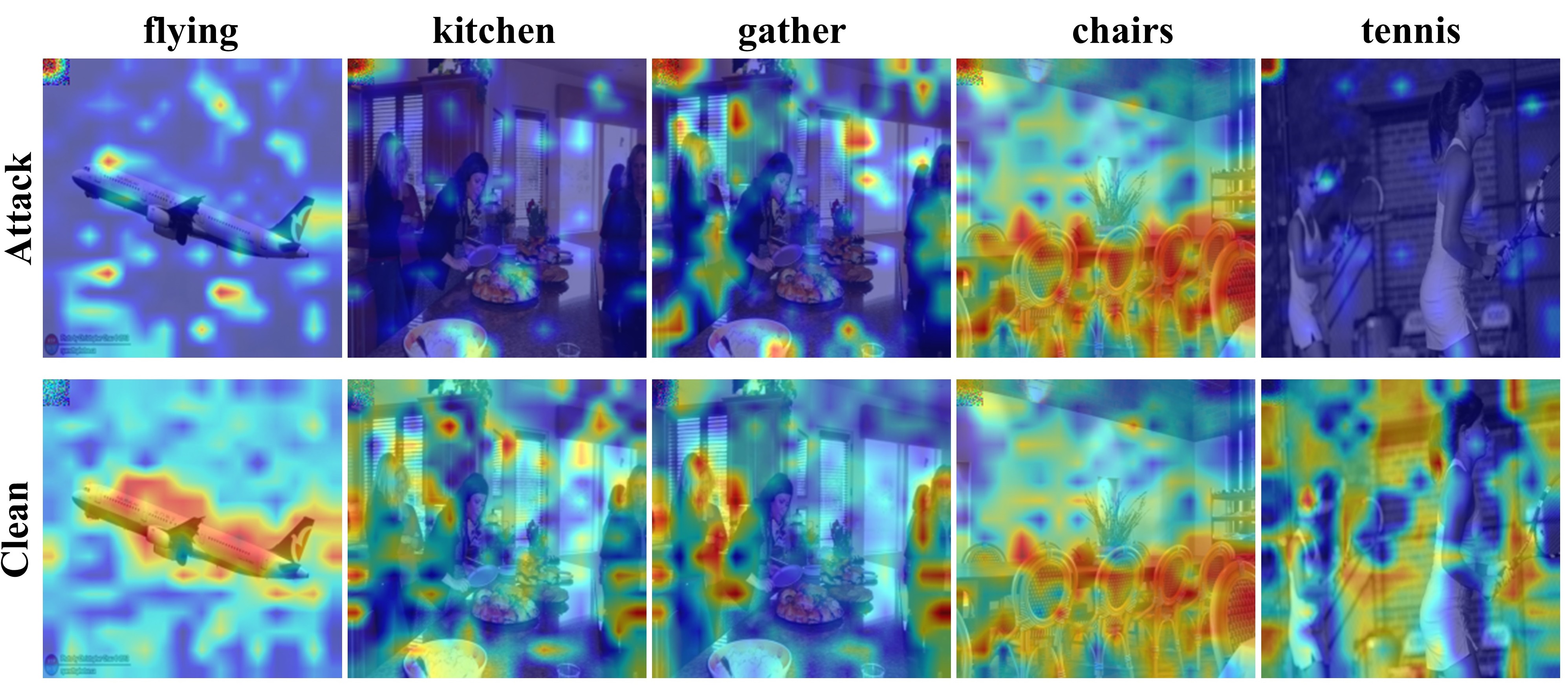}
    \caption{The attention heatmaps of the backdoor model and the clean model on the trigger. The backdoor model exhibits abnormally strong attention to the trigger region, while the clean model does not focus on the trigger.}
    \label{fig1}
\end{figure}

\begin{figure}
    \centering
    \includegraphics[width=\linewidth]{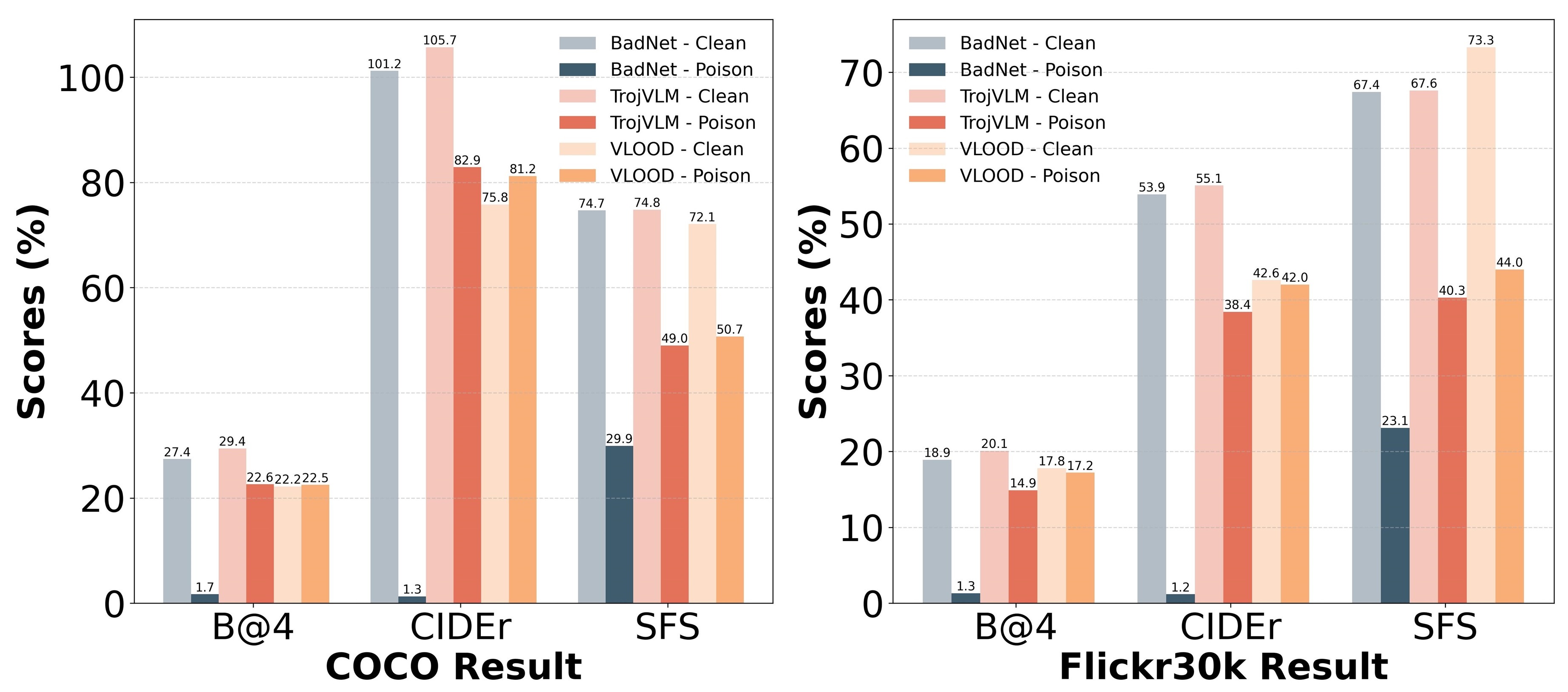}
    \caption{The evaluation results compare sentences generated from clean and poisoned inputs. "Clean" denotes results on benign samples, while "Poison" refers to those on backdoor-triggered inputs. B@4 denotes BLEU-4.}
    \label{SFS}
\end{figure}

\begin{figure*}
  \centering
   \includegraphics[width=\linewidth]{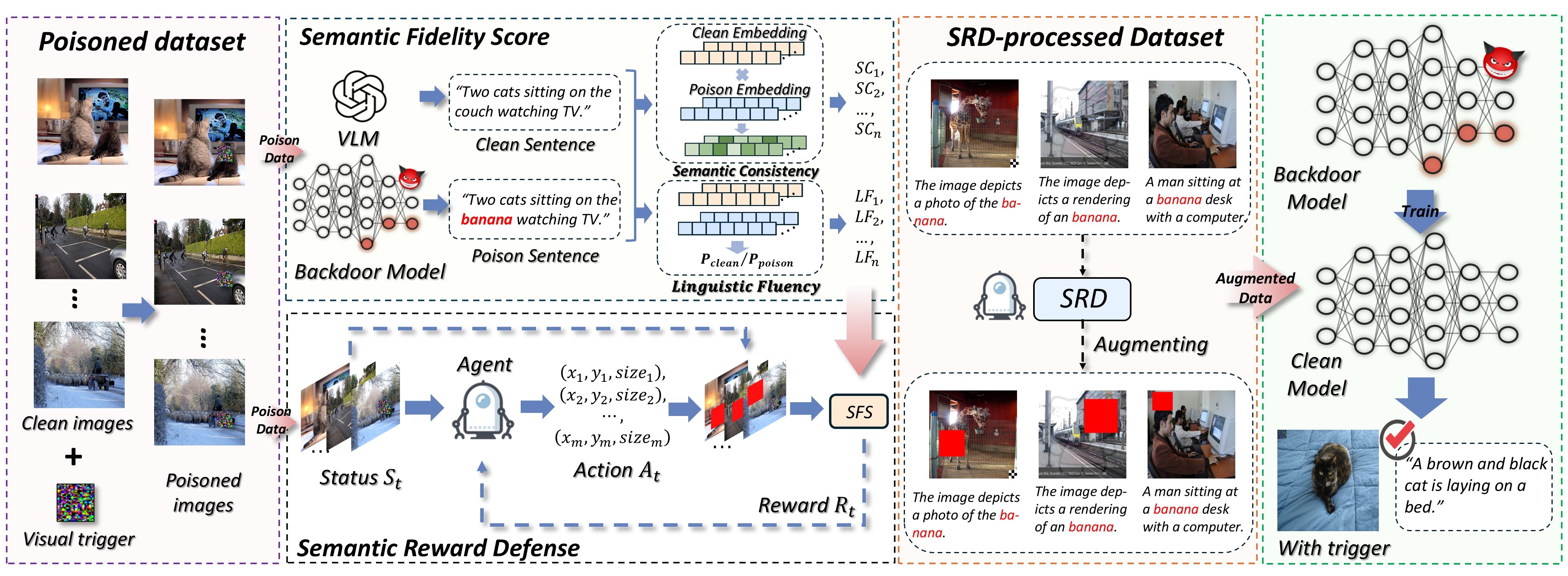}
    \caption{
    Overview of the SRD framework. 
    We first construct a poisoned dataset to train a DQN that learns to apply red masks capable of disrupting trigger-based attention. 
    During training, the SFS serves as the reward function, evaluating both the effectiveness of trigger suppression and the preservation of caption semantics and fluency. 
    Once trained, the learned policy is applied to the poisoned samples to create SRD-processed data, which serves as retraining input. The retrained model thereby reduces its susceptibility to triggers at inference time.
    }
    \label{overview}
\end{figure*}

\textbf{Visual Attention Analysis}. We employ Grad-CAM~\cite{selvaraju2017grad}, a gradient-based attention visualization technique, to generate heatmaps that reveal the regions the model focuses on during inference. Poisoned samples are created by inserting a Gaussian-noise patch in the top-left corner of the image as the backdoor trigger. Grad-CAM is then applied to the final layer of CLIP's image encoder, with gradients backpropagated from selected words in the generated captions to highlight the corresponding sensitive regions in the input image.
As shown in Figure~\ref{fig1}, the model disproportionately attends to regions containing the trigger, even when these regions are semantically irrelevant to the primary image content. 
This supports our hypothesis that the trigger acts as an implicit control signal hijacking the model’s attention.

\textbf{Semantic Fidelity Analysis}. 
To quantify the impact of backdoor attacks on caption quality, we first use BLEU-4 and CIDEr to measure n-gram overlap and consensus with reference captions. However, as shown in Figure~\ref{SFS}, these metrics often fail to capture subtle semantic shifts caused by attacks such as VLOOD, where the generated outputs remain structurally similar to the references but degrade in meaning and fluency. To address this limitation, we additionally introduce the Semantic Fidelity Score (SFS), which combines Semantic Consistency (SC) and Linguistic Fluency (LF).

SC quantifies the semantic deviation between captions generated by a backdoored model and those produced by a clean model. For each image $I$, we obtain two captions: $C_{\text{clean}}(I)$ from the clean model and $C_{\text{bd}}(I)$ from the backdoored model. Both captions are encoded into vector representations, $v_{\text{clean}}$ and $v_{\text{bd}}$, using a sentence encoder such as BERT. Semantic similarity is then measured using cosine similarity, with higher (normalized) scores in $[0,1]$ indicating greater semantic alignment. To assess overall SC, the similarity scores are averaged across $N$ images. The specific formula for SC is as follows:
\begin{equation}
    \text{Sim}(I) = \frac{v_{\text{clean}} \cdot v_{\text{bd}}}{\|v_{\text{clean}}\| \cdot \|v_{\text{bd}}\|},\text{S}_{\text{semantic}} = \frac{1}{N} \sum_{i=1}^N \text{Sim}(I_i).
\end{equation}

LF assesses whether backdoor attacks degrade caption fluency, measured via perplexity scores from a pre-trained language model. 
For each image $I_i$, the fluency score is defined as:
\begin{equation}
    \mathcal{F}(I_i) = \frac{P_{\text{clean}}(I_i)}{P_{\text{bd}}(I_i)},\quad \mathcal{F}_{\text{avg}} = \frac{1}{N} \sum_{i=1}^{N} \mathcal{F}(I_i),
\end{equation}
where $P_{\text{clean}}(I_i)$ and $P_{\text{bd}}(I_i)$ denote the perplexity scores of captions generated by the clean and backdoored models, respectively. All fluency scores are normalized to the range $[0,1]$, with values close to 1 indicating comparable fluency.

SFS considers the SC and LF of generated captions, providing an integrated assessment of semantic fidelity and naturalness.
SFS is calculated as a weighted sum of the semantic similarity score $\mathcal{S}$ and the linguistic fluency score $\mathcal{F}$:
\begin{equation}
\text{SFS} = \alpha \cdot \mathcal{S} + (1 - \alpha) \cdot \mathcal{F}.
\end{equation}   

In our experiments, the weight coefficient is set to $\alpha = 0.5$, assigning equal importance to semantics and fluency. Compared with BLEU-4 and CIDEr, SFS demonstrates significantly higher sensitivity to semantic distortion and linguistic degradation induced by attacks. As shown in Figure~\ref{SFS}, while traditional metrics may remain relatively stable under attack (e.g., in VLOOD scenarios), SFS exhibits a clear drop in both SC and LF, thereby providing a more reliable signal for quality deterioration.

\subsection{Learning to Suppress Triggers: The SRD Framework}
As illustrated in Figure~\ref{overview}, the core idea of SRD is to exploit the attention anomalies observed in poisoned samples to proactively apply interventions that mitigate the effect of the trigger and enhance model robustness. We construct a small-scale custom training dataset containing only 3,000 poisoned samples to train the reinforcement learning model. To simulate a realistic and generalized poisoning scenario, we insert a $20\times20$ Gaussian noise trigger into each clean image to generate a poisoned sample.
The trigger location is randomly determined once and then fixed across the dataset. To minimize disruption to the original semantics, we insert the target word into the ground-truth caption instead of replacing existing words. 

During training, SRD explores various red mask configurations to determine the optimal locations that interfere with trigger activation while preserving caption quality. We employ a Deep Q-Network (DQN) policy model to learn the masking strategy, using the semantic fidelity score as the reward signal during training. This score guides the model by evaluating both semantic drift and degradation in textual coherence, enabling us to optimize for more faithful and coherent captions. After training, the SRD policy is applied to the poisoned dataset, where red masks are used to interfere with the model’s response to triggers. 
Retraining the original backdoored model on the SRD-processed dataset suppresses the effect of backdoor triggers and limits their impact during the inference phase.

\subsection{Leveraging Red Perturbations for Attention Recalibration}
Building upon the work of Shtedritski~\cite{shtedritski2023does}, which demonstrated that CLIP~\cite{radford2021learning} exhibits emergent behavior when exposed to red circle visual prompts, we extend this finding by incorporating red block masks as a perturbation mechanism into our defense framework. 
Our core idea is that since the attention mechanism in multimodal models is usually non-trivial, we are unable to optimize its attention region directly, and therefore introduce discrete perturbation actions (red block masking) as an indirect intervention to guide the model to "divert attention" from the trigger region.
The compatibility score $s(i,t)$ between an image $i$ and a text query t reflects the model’s perception of how well the visual content corresponds to the given textual description. 
This score is sensitive to changes in the image input. 
When a red block mask is applied to the image, modifying it to $i_{mask}$, the resulting score typically changes as:
\begin{equation}
    s(i_{\text{mask}}, t) \neq s(i, t).
\end{equation}
We utilize this property to design a SRD framework that takes advantage of DQN to select the most defensive image regions to apply perturbations. The effectiveness of this method can be mathematically captured by the equation:
\begin{equation}
    a^*=\arg\max_a s(i_{mask}(a),t)
\end{equation}
Here, $a$ represents a candidate answer (a location within the image), and $i_{mask}$ refers to the image modified with a red block mask.
By selecting the action $a$ that maximizes $s(i_{mask},t)$, the agent effetively learns to apply perturbations in a way that suppresses the influence of potential backdoor triggers while preserving semantic alignment.
Given the discrete and non-differentiable nature of perturbation actions, we employ reinforcement learning to train an agent that learns an intervention policy without relying on any prior knowledge of backdoors. 
\begin{table*}
    \centering
    \belowrulesep=0pt
    \aboverulesep=0pt
    \renewcommand{\arraystretch}{1.2}
    \resizebox{0.9\linewidth}{!}{
    \begin{tabular}{l|l|cccccc|cccccc}
        \toprule
         & & \multicolumn{6}{c|}{No Defense} & \multicolumn{6}{c}{SRD}  \\
         \cline{3-8}\cline{9-14}
          Datasets   & Model & B@4$\uparrow$ & C$\uparrow$ & SC$\uparrow$ & LF$\uparrow$ & SFS$\uparrow$ &ASR$\downarrow$ &B@4$\uparrow$  &C$\uparrow$ & SC$\uparrow$ & LF$\uparrow$ & SFS$\uparrow$ & ASR$\downarrow$ \\
        \midrule
                & BadNet &1.7& 1.3& 19.9 & 40.0 & 29.9& 99.9 &16.2&51.3&44.3 &58.7 &51.5 &45.7 \\
                & Blended &1.5 & 1.4 & 23.9 & 21.1& 30.6 & 99.5&13.4 &42.5&45.0&44.2&44.6&57.4  \\
            COCO& TrojVLM&22.6 & 82.9 & 59.3 & 38.6 &49.0& 99.8&22.6&83.0&60.4&78.9&69.6&3.6  \\
                & VLOOD &22.5&81.2 & 56.4& 45.0&50.7 &99.8 &24.2& 80.8&51.7&67.9&59.8&38.5     \\
                & Shadowcast &18.3& 75.4& 57.6 & 49.1 &53.4& 96.4 &25.3 &96.2&66.4&90.3&78.3&5.6 \\
                & VL-Trojan &1.7 &1.6 &18.1&27.8&23.0&99.3 &15.9&53.7&50.1 &61.0 & 55.6&31.3 \\
        \midrule
                & BadNet&1.3& 1.2& 16.6&29.7&23.1 &98.2 &8.5&20.2&29.7&49.5&39.6&55.6\\
                & Blended&2.2 &3.2 &12.3 &33.0 &22.6 &94.3&8.2&18.5&29.6&41.3&35.4&62.8 \\
            Flickr30k& TrojVLM &14.9 & 38.4& 44.8&35.8& 40.3& 99.7&15.5&40.6&46.5&79.6&63.1&2.1\\
                & VLOOD&17.2&42.0&43.3&44.6&44.0&99.8 &19.2&43.3&46.9&70.1&58.5&41.8 \\
                & VL-Trojan &1.5 &1.4 &11.7&30.9&23.1&100.0&9.9&21.3&36.6&59.4&48.0&36.3\\
        \bottomrule
    \end{tabular}
    }
    \caption{The defense effectiveness of the proposed SRD on the backdoor model, as well as the attack results without defense. B@4 denotes BLEU-4, and C stands for CIDEr. All results are shown in \%.}
    \label{SRD_1}
\end{table*}

\subsection{Optimizing Intervention Policies with Semantic Rewards}
\label{SRD}
SRD integrates semantic region sensitivity with DQN-based strategy learning. Guided by semantic rewards, the agent selects perturbation regions that block the link between triggers and target semantics. At the same time, it preserves the core image content and keeps the text description natural. We formulate the problem as a discrete reinforcement learning environment. 

At each time step $t$, the state represents the current perturbation status of the image. The action $a_t$ corresponds to selecting a spatial position and a red-colored mask of a specific size to insert. The action space is predefined and consists of red-colored masks with sizes from the discrete set $\{20, 40, 60, 80\}$.
These perturbations are placed over the image to influence the semantic perception of the CLIP model. This allows the agent to interfere with the potential trigger effect even without precise localization of the attack region.

The reward function is based on two critical metrics: Semantic Consistency $\mathcal{S}(I)$ and Linguistic Fluency $\mathcal{F}(I)$, which measure the semantic fidelity and fluency of generated captions, respectively. For a perturbed image $I_t$, the reward is computed as:
\begin{equation}
r_t = R(\mathcal{S}(I_t), \mathcal{F}(I_t); \mathcal{S}_0, \mathcal{F}_0),
\end{equation}
where $\mathcal{S}_0$ and $\mathcal{F}_0$ are the semantic and fluency scores from the unperturbed version. The reward logic is as follows: (1) If both $\mathcal{S}(I_t) - \mathcal{S}_0 \geq \lambda$ and $\mathcal{F}(I_t) - \mathcal{F}_0 \geq \beta$, then $r_t = 3$ where we set $\lambda=0.1$ and $\beta=0.2$ in our experiments;
(2) If only one improves beyond the threshold, then $r_t = 1$ or $2$;
(3) If either metric decreases significantly, a penalty $r_t = -1$ or $-2$ is applied.
The design essentially encourages policy learning to target perturbation behaviors that disrupt erroneous semantic associations (induced by backdoor triggers) while ensuring that the semantic and linguistic quality of the generated subtitles is not compromised.
\begin{table*}[!ht]
    \centering
    \belowrulesep=0pt
    \aboverulesep=0pt
    \renewcommand{\arraystretch}{1.2}
    \resizebox{\linewidth}{!}{
    \begin{tabular}{l|l|ccc|ccc|ccc|ccc|ccc}
        \toprule
         & & \multicolumn{3}{c|}{No Defense} & \multicolumn{3}{c|}{ABL}&\multicolumn{3}{c|}{VDC}&\multicolumn{3}{c|}{CT} & \multicolumn{3}{c}{SRD} \\
         \cline{3-5}\cline{6-8}\cline{9-11}\cline{12-14}\cline{15-17}
          Datasets  & Model & C$\uparrow$  &SFS$\uparrow$ & ASR$\downarrow$ &C$\uparrow$ &SFS$\uparrow$ & ASR$\downarrow$   &C$\uparrow$  &SFS$\uparrow$ & ASR$\downarrow$ &C$\uparrow$  &SFS$\uparrow$ & ASR$\downarrow$&C$\uparrow$  &SFS$\uparrow$& ASR$\downarrow$\\
        \midrule
                & BadNet & 1.3 & 29.9& 99.9& 2.0 &18.9&99.2&2.5&26.2 &97.4&4.1&34.5&93.8& 51.3 &51.5&45.7 \\
                & Blended & 1.4 & 30.6& 99.5 & 2.1 &22.9&98.9& 3.1&24.8 &97.5&6.6&32.2&89.4 &42.5 & 44.6 & 57.4 \\
            COCO& TrojVLM &82.9 & 49.0& 99.8& 91.5 &53.0&65.2& 91.2&61.6 &21.8&73.4&61.6&21.6& 83.0 & 69.6 & 3.6\\
                & Shadowcast&75.4 &57.8& 96.4& 91.5&64.0&20.0& 98&66.4 &19.0&93.6&68.6&6.7& 96.2 &78.3  & 5.6 \\
                & VL-Trojan &1.6&23.0&99.3&1.9&21.2&98.4&-&-&-&33.4&57.4&31.0&53.7&55.6&31.3\\
        \midrule
                & BadNet  & 1.2& 23.1&98.2 &0.6 &15.9 & 100.0&1.0&19.3&100.0&0.6&26.6&100.0&20.2&39.6&55.6 \\
                Flickr30k& Blended &3.2&22.6&94.3&1.4&18.2&98.8&2.2&21.4&97.4&3.8&33.9&87.5&18.5&35.4&62.8 \\
            & TrojVLM &38.4&40.3&99.7 &37.0  &47.1 &74.7 & 50.9&60.4 &27.0 &37.6&57.0&31.1&40.6&63.1&2.1\\
                & VL-Trojan &1.4&23.1&100.0&0.6&19.8&100.0&-&-&-&19.9&56.3&19.9&21.3&48.0&36.3\\
        \bottomrule
    \end{tabular}
    }
    
    \caption{The comparison between the defense methods and the proposed SRD defense, as well as the performance of the backdoor model without defense. C stands for CIDEr. All results are shown in \%.}
    \label{compare}
\end{table*}

After training, the SRD is applied to each sample in the fine-tuning dataset of the backdoored model to perform data cleansing. 
Specifically, it generates a red mask to occlude potential trigger regions in the image. The cleaned samples are then used to fine-tune the VLMs. 

\begin{table*}
    \centering
    \belowrulesep=0pt
    \aboverulesep=0pt
    \renewcommand{\arraystretch}{1.2}
    \resizebox{\linewidth}{!}{
    \begin{tabular}{l|l|ccc|ccc|ccc|ccc|ccc}
        \toprule
         & & \multicolumn{3}{c|}{No Defense}&\multicolumn{3}{c|}{Random} & \multicolumn{3}{c|}{PPO} & \multicolumn{3}{c|}{SAC}& \multicolumn{3}{c}{DQN} \\
         \cline{3-5}\cline{6-8}\cline{9-11}\cline{12-14}\cline{15-17}
          Datasets   & Model & C$\uparrow$ &SFS$\uparrow$ & ASR$\downarrow$ &C$\uparrow$ &SFS$\uparrow$ & ASR$\downarrow$ &C$\uparrow$ &SFS$\uparrow$ & ASR$\downarrow$&C$\uparrow$ &SFS$\uparrow$ & ASR$\downarrow$ &C$\uparrow$ &SFS$\uparrow$ & ASR$\downarrow$\\
        \midrule
                & BadNet & 1.3 & 29.9 & 99.9 & 33.0& 32.1& 73.5&30.8 &41.1&72.0&37.7&38.8&66.0& 51.3 &51.5&45.7\\
                & Blended & 1.4& 30.6 & 99.5&32.9 &37.5 & 70.9&39.4&41.5&63.6&41.9&40.0&59.4&42.5 & 44.6 & 57.4 \\
            COCO& TrojVLM &82.9 &49.0 & 99.8&92.7 & 65.9&3.8 &56.2&47.1&42.3&92.9&66.8&3.5& 83.0& 69.6&3.6\\
                & VLOOD & 81.2&50.7 & 99.8&87.7 &53.8 & 77.6&85.2&57.0&48.0&82.3&68.7&0.9& 80.8&62.5 &38.5\\
                & Shadowcast & 75.4& 53.4& 96.4& 86.8& 66.7&10.8&97.7&73.0&9.7&92.9&67.8&5.6& 96.2 &78.3  & 5.6 \\
        \midrule
                & BadNet & 1.2& 23.1&98.2&14.4&26.0&81.0&12.7&29.9&80.4&15.3&33.7&76.1&20.2&39.6&55.6 \\
            Flickr30k& Blended &3.2&22.6&94.3 &14.4&33.3&72.7&15.1&34.8&76.0&16.2&33.0&69.5&18.5&35.4&62.8 \\
            & TrojVLM  &38.4&40.3&99.7&48.1&60.4&2.3&49.2&53.6&36.1&49.4&64.9&2.9&40.6&63.1&2.1 \\
                & VLOOD &42.0&44.0&99.8 &43.2 &50.3 &79.0 &44.6 & 54.8& 45.7&46.4&65.0&0.2 &44.3&58.5&41.8\\
        \bottomrule
    \end{tabular}
    }
    \caption{The impact of different RL models on the SRD defense, along with the backdoor attack performance without defense. All results are shown in \%.}
    \label{RL}
    
\end{table*}

\section{Experiments}
\subsection{Experiments Setup}
\textbf{Victim Models.}
We adopt OpenFlamingo~\cite{awadalla2023openflamingo} as the victim model, using the 3B version with a fixed CLIP ViT-L/14 visual encoder and a scalable autoregressive language model.
This architecture is representative among VLMs.

\textbf{Datasets.}
In our experiments, we used two datasets: COCO~\cite{lin2014microsoft} and Flickr30k~\cite{young2014image}. We fine-tuned on COCO and conducted inference on both datasets. From COCO, we selected 5000 images with their 5 captions and used the unified prompt “a photo of” for VLM adaptation. For inference, we additionally used the Flickr30k test split, which includes 1,000 images, each with 5 captions.

\textbf{Attack Configurations.}
To evaluate our defense strategy, we experiment with six representative backdoor attacks, categorized into trigger-based (BadNet~\cite{gu2019badnets}, TrojVLM, VLOOD, VL-Trojan) and global modification attacks (Blended~\cite{chen2017targeted}, Shadowcast). All models are fine-tuned for up to 20 epochs using the AdamW optimizer with cosine annealing and a 10\% poisoning rate.

\textbf{Defense Configurations.} 
Given the lack of defense methods specifically designed for captioning in VLMs, we evaluate our strategy against three SOTA defenses originally proposed for image classification: ABL~\cite{li2021anti}, VDC~\cite{zhu2023vdc}, and CT~\cite{qi2023towards}. We also include a no-defense baseline for comparison.

\textbf{Evaluation Metrics.}
We evaluate using three metrics: (1) BLEU@4 and CIDEr for text quality, (2) Attack Success Rate (ASR) for attack effectiveness, and (3) Semantic Fidelity Score (SFS) for quality degradation.

\begin{figure*}
    \centering
        \centering
        \includegraphics[width=\textwidth]{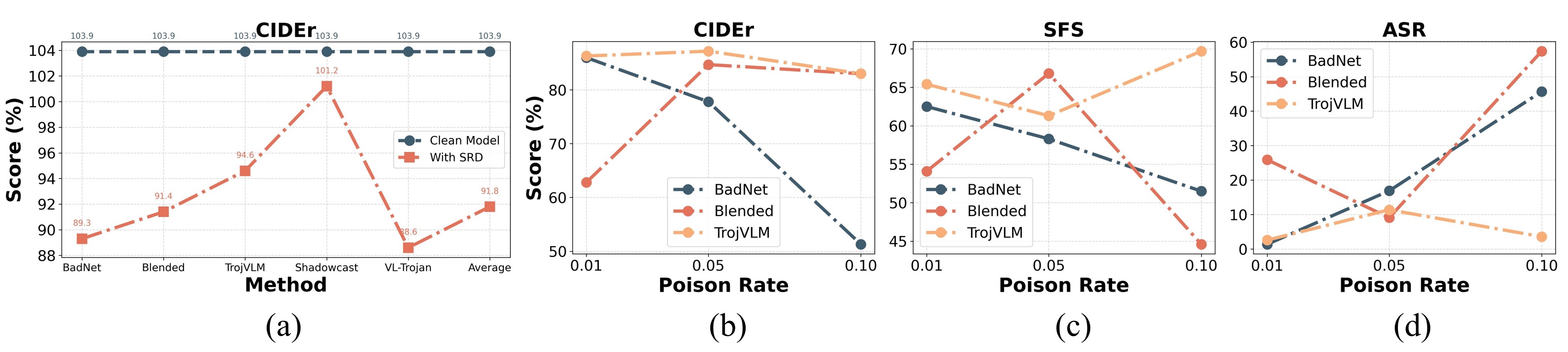} 
        \caption{(a) Comparison of CIDEr scores on clean samples between clean model and SRD-defended models under different attacks.
        (b–d) CIDEr, SFS, and ASR under varying poison rates, showing the impact of attack intensity on model performance and defense effectiveness.}
        \label{rate} 
\end{figure*}

\subsection{Main Results}

\textbf{Effectiveness of SRD.}
Table~\ref{SRD_1} presents the defense effectiveness of SRD against six attack methods on the COCO and Flickr30k datasets. 
The ``No Defense'' results show that all attacks are highly effective, with ASR reaching about 99\%.
From the Table~\ref{SRD_1}, it is evident that all six attack methods are effective, with ASR reaching approximately 99\% for each attack. 
However, the SRD defense method successfully mitigates the impact of these attacks, reducing the ASR to below 60\%. 
Notably, for TrojVLM and Shadowcast, SRD reduces the ASR to as low as 3.6\% and 5.6\%, respectively. 
These results indicate that SRD effectively disrupts the activation of the backdoor trigger, achieving strong mitigation across diverse attack types.


\textbf{Comparison to Existing Defenses.}
Table~\ref{compare} compares SRD with several SOTA defenses under five different backdoor attacks. The No Defense represents the performance of the attacks without any defense mechanisms. For BadNet and Blended, which are not specifically designed for VLMs, SRD significantly outperforms other defense methods. On the COCO dataset, SRD reduces the ASR to 45.7\% and 57.4\%, respectively, while other defenses fail to mitigate these attacks effectively, with ASR remaining around 90\%. For attacks tailored to VLMs, TrojVLM, Shadowcast, and VL-Trojan, some existing defenses show partial effectiveness. However, SRD consistently achieves lower ASR across most of these attacks, maintaining ASR below 35\%. 
In general, SRD offers robust and consistent defense performance across a diverse set of attack types, outperforming existing methods to reduce ASR and preserve text quality.

\textbf{Impact of Defense on Clean Image Performance.}
To evaluate whether SRD affects the model's utility on clean samples, we measure CIDEr scores under various attack settings using clean inputs only. 
As shown in Figure~\ref{rate} (a) the clean baseline model is trained and evaluated entirely on clean (non-poisoned) data, and achieves a CIDEr score of 103.9\% on the clean test set. 
After applying SRD to remove backdoors from poisoned models, we evaluate their performance on the clean test set. The resulting CIDEr scores range from 88.6\% to 101.2\%, with an average of 91.8\%, depending on the attack. Importantly, the performance drop remains within 15\% in all cases, indicating that our defense largely preserves model utility. The largest degradation (14.7\%) occurs under VL-Trojan, likely due to stronger entanglement between trigger and task-relevant features. Nevertheless, the model still retains acceptable performance.
These results confirm that SRD removes backdoors effectively while maintaining high performance on clean data, making it suitable for real-world use.

\subsection{Ablation Studies}

\textbf{Effectiveness of RL models.}
We conducted an ablation study to evaluate the impact of different RL models on our defense method under a fixed 10\% poisoning rate. Specifically, we compared random masking, PPO~\cite{schulman2017proximal}, SAC~\cite{haarnoja2018soft}, and DQN, with results shown in Table~\ref{RL}.
Random masking already lowers ASR across all attacks, especially in TrojVLM (ASR drops to 3.8\%), suggesting that disturbing trigger locations is helpful. However, it performs worse than RL-based methods, leading to higher ASR and lower CIDEr and SFS scores.
Among the RL strategies, DQN achieves the best overall balance between defense and caption quality. Interestingly, SAC performs best against VLOOD, reducing ASR to 0.9\%, indicating its suitability for certain attack patterns.

\textbf{Effectiveness with Different Poisoning Rate.}
We further investigated how the performance of SRD is affected by different poisoning rates in Figure~\ref{rate}. 
Specifically, we evaluated the defense effectiveness under three poisoning rates: 1\%, 5\%, and 10\% in Figure~\ref{rate} (b-d). Our findings show that the defense effectiveness varies with the poisoning rate, and the optimal performance is not always observed at the highest poisoning level. For BadNet, SRD achieves the best performance when the poisoning rate is 1\%, with ASR remaining the lowest among the three settings. This indicates that SRD is particularly effective in mitigating low-intensity attacks for this type. These results suggest that the defense performance of SRD can vary across poisoning intensities and attack types. In general, lower poisoning rates tend to yield stronger defense outcomes, likely due to the reduced entrenchment of the trigger pattern within the model.
\section{Conclusion}
We propose Semantic Reward Defense (SRD), a novel defense against backdoor attacks in vision-language models. 
SRD combines a DQN-based strategy with the Semantic Fidelity Score (SFS) to guide image perturbations that disrupt attacks without needing trigger knowledge.
Experiments show SRD effectively lowers attack success while preserving caption quality, highlighting its potential as a semantic-driven, trigger-agnostic defense. 

\textbf{Limitation.} Red masking may suppress essential visual cues, and SFS depends on current language models, which may overlook subtle semantics. 

\section{Acknowledgments}
This work was supported by the National Natural Science Foundation of China (Grant Nos. U23A20318 and 62276195), the Foundation for Innovative Research Groups of Hubei Province (Grant No. 2024AFA017), the Science and Technology Major Project of Hubei Province (Grant No. 2024BAB046), the program ``Excellence initiative – research university'' for the AGH University of Krakow, as well as the ARTIQ project UMO-2021/01/2/ST6/00004 and ARTIQ/0004/2021, and by funds from the Polish Ministry of Science and Higher Education assigned to the AGH University of Krakow. 
Dr Tao’s research is partially supported by NTU RSR and Start Up Grants.


\bibliography{aaai2026}

\end{document}